\documentclass[conference]{IEEEtran}
\IEEEoverridecommandlockouts
\usepackage{cite}
\usepackage{amsmath,amssymb,amsfonts}
\usepackage{algorithmic}
\usepackage{graphicx}
\usepackage{textcomp}
\usepackage{xcolor}
\usepackage{booktabs}
\usepackage{multirow}
\def\BibTeX{{\rm B\kern-.05em{\sc i\kern-.025em b}\kern-.08em
    T\kern-.1667em\lower.7ex\hbox{E}\kern-.125emX}}

\begin{document}

\title{A Novel Unified Approach to Deepfake Detection}

\author{\IEEEauthorblockN{Lord Sen}
\IEEEauthorblockA{\textit{Computer Science and Engineering} \\
\textit{NIT Rourkela}\\
Rourkela, Odisha\\
lordsen3008@gmail.com}
\and
\IEEEauthorblockN{Shyamapada Mukherjee}
\IEEEauthorblockA{\textit{Computer Science and Engineering} \\
\textit{NIT Rourkela}\\
Rourkela, Odisha \\
mukherjees@nitrkl.ac.in}
}

\maketitle

\begin{abstract}
Advancements in the field of AI are increasingly giving rise to various threats, one of the most prominent being the synthesis and misuse of deepfakes. To sustain trust in this digital age, the detection and tagging of deepfakes are essential. In this paper, a novel architecture for deepfake detection in images and videos is presented. The architecture uses cross-attention between spatial and frequency domain features to classify an image as real or fake. This paper aims to develop a unified architecture and provide insights into each step. Through this approach, we achieve results competitive with the State-of-the-Art (SOTA). Furthermore, this approach generalizes well to cross-dataset testing, which is evident from the experimental results.
\end{abstract}


\section{Introduction}
Deepfake media is becoming an increasing threat in today’s digital world, where trust in what we see and hear online is decreasing rapidly. These AI-generated fake videos, images, and audio can create realistic but entirely false depictions of people. This technology can be misused to spread lies about public figures, manipulate opinions, or even ruin someone’s reputation. As they become easier to create and harder to detect, they challenge our ability to tell what is real from what is not. The rise of deepfakes highlights how important it is to develop tools to identify them and to raise awareness about their potential impact on trust and truth in the digital age.

To detect deepfakes we first need to understand its generation. Deepfake generation can be categorized into 4 mainstream research areas:

\textbf{Face Swapping:} This task involves replacing the identity of a target face with that of a source face while preserving ID-irrelevant attributes like lighting and expressions.

\textbf{Face Reenactment:} This technology focuses on transferring facial movements and head poses from a driving source to a target image without altering the target's identity.

\textbf{Talking Face Generation: }This area aims to naturally match a character's lip and mouth movements to driving textual or audio content.

\textbf{Facial Attribute Editing:} This process modifies specific semantic features of a target face, such as age, skin tone, or expression. 

Having a brief idea on the various streams of deepfake generation, let's dive into the various types of models or architectures used for deepfake generation.

\textbf{Pre GAN Generators: }Before the widespread adoption of adversarial training, early face-swapping pipelines relied heavily on Variational Autoencoder (VAE) architectures to alter the relationship between latent features and linear mappings. These foundational models often utilized dual encoders to process different facial regions separately, explicitly disentangling identity features from attribute and expression information. By representing the face and other elements in independent latent spaces, these early architectures enabled the direct replacement of identity information through latent space interpolation.

\textbf{GANs as Deepfake Generators: }The synthesis of highly realistic deepfakes has been fundamentally driven by advancements in Generative Adversarial Networks (GANs) and Variational Autoencoders (VAEs). GANs generate convincing synthetic media through an adversarial process—a continuous competition between a generator and a discriminator—that steadily improves the visual fidelity of the outputs. Various architectures have been specialized to handle distinct manipulation tasks. For instance, models like Pix2Pix specialize in image-to-image translation frameworks, which are essential for altering specific visual elements. To achieve higher resolution and precision, StyleGAN introduces style-based controls for the fine-grained manipulation of features like facial attributes, while StyleGAN2 further stabilizes the training process to enhance both overall quality and controllability. Additionally, hybrid approaches such as CVAE-GAN integrate VAE-based representations with adversarial learning to enrich the model's latent feature modeling capabilities.

\textbf{Diffusion Models as Deepfake Generators: }Modern generative frameworks are increasingly dominated by Denoising Diffusion Probabilistic Models (DDPMs) and Latent Diffusion Models (LDMs) (e.g., Stable Diffusion, Midjourney, Sora). By modeling generation as an iterative reverse-diffusion process across latent manifolds, diffusion models largely bypass the mode collapse and training instability inherent to GANs. Consequently, they generate media with unprecedented structural coherence, photorealistic high-frequency textures, and fewer obvious spatial artifacts.

\textbf{Transformers and ViTs as Deepfake Generators: }Furthermore, deep Convolutional Neural Networks (CNNs) frequently serve to manipulate localized facial features, whereas recent transformer-based architectures capture global, long-range dependencies within images to produce highly convincing deepfakes.

These techniques, combined with increasing computational power, make deepfake images increasingly difficult to distinguish from real ones. Because deepfakes pose significant risks like privacy invasion and the spread of harmful content, developing robust forgery detection is critical. Important domains of deepfake detection are: 

\textbf{Spatial Domain Methods:} These approaches identify tampering by analyzing image-level inconsistencies, boundary artifacts, and local noise anomalies.

\textbf{Temporal Domain Methods:} These techniques detect forged videos by evaluating unnatural physiological features, like irregular blinking, or inter-frame inconsistencies.

\textbf{Frequency Domain Methods:} These systems focus on identifying mid- and high-frequency spectrum flaws or periodic artifacts left behind during the deepfake generation process.

Section II of this paper briefly reviews current methods for deepfake detection in images, videos, and audio. In Section III, the proposed architecture is discussed. The results obtained and the conclusions drawn, along with future plans, are discussed in Sections IV and V.
\section{Literature Review}
In \cite{b2}, a modified InceptionResNetV2 model based on the Constant-Q Transform (CQT) method to extract deep time-frequency features from audio has been suggested, and it is combined with an XceptionNet model for video frames to detect deepfakes in videos. Architectures to detect deepfakes using CNNs are presented in \cite{b3} and \cite{b10}, audio analysis in \cite{b11}, cross-domain analysis in \cite{b14}--\cite{b20}, vision transformers in \cite{wodajo2021deepfake, zhuang2022uia, b61}, and Fourier transforms in \cite{b8}. A live skin detection system is proposed in \cite{b5}. Furthermore, \cite{b20} presents a trend of the mean and correlation coefficient of real and fake images in the frequency domain. In this paper, we identify underlying trends of energy, entropy, and power spectral density (PSD). This paper is an extension of our previous conference paper~\cite{sen2024novel}, where we had tested on smaller subsets of FF++ dataset with just 400 images. In this version, we have added comparisons with SOTA methods and removed the Blood Detection Module.
\section{Frequency Domain Analysis}

\subsection{Analysis of Global Spatial Intensity Moments}

\begin{figure*}[htbp]
    \centering
    \includegraphics[width=\linewidth]{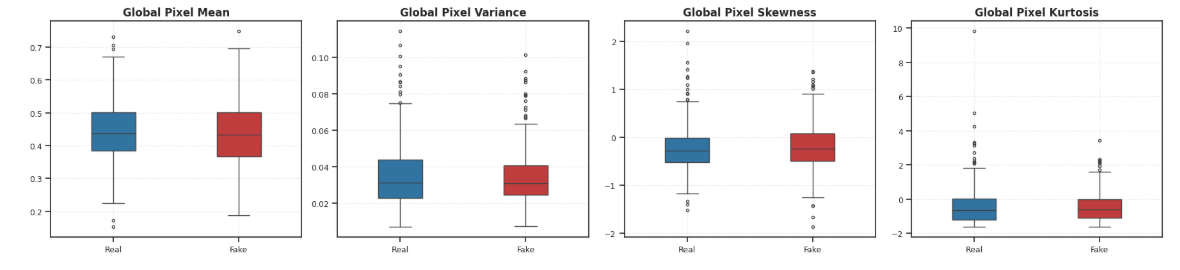}
    \caption{Distributions of global spatial intensity moments across the dataset for authentic (blue) and manipulated (red) faces: Global Pixel Mean, Variance, Skewness, and Kurtosis.}
    \label{fig:global_moments}
\end{figure*}

To evaluate whether raw pixel-level intensity statistics can provide trivial cues for forgery detection prior to frequency transformation, we compute the first four standardized statistical moments—mean, variance, skewness, and excess kurtosis—over the normalized grayscale pixel intensities ($I \in [0, 1]$) of the cropped faces across the entire dataset (Fig.~\ref{fig:global_moments}):
\[
\mu = \frac{1}{N}\sum_{i=1}^N I_i, \quad \sigma^2 = \frac{1}{N}\sum_{i=1}^N (I_i - \mu)^2
\]
\[
\gamma_1 = \frac{1}{N}\sum_{i=1}^N \left(\frac{I_i - \mu}{\sigma}\right)^3, \quad \gamma_2 = \frac{1}{N}\sum_{i=1}^N \left(\frac{I_i - \mu}{\sigma}\right)^4 - 3
\]
where $N = H \times W$ denotes the total pixel count.

The distribution profiles across both cohorts highlight several key characteristics:
\begin{itemize}
    \item \textbf{Photometric Consistency (Mean and Variance):} The median global pixel intensity ($\approx 0.44$) and intra-frame variance ($\approx 0.031$) are virtually identical between authentic and synthesized faces. This confirms that modern deepfake generation pipelines employ color-matching, histogram matching, and Poisson blending steps, eliminating macro-level luminance and contrast discrepancies relative to the target video frames.
    \item \textbf{Tonal Distribution Symmetry (Skewness):} Both classes exhibit closely matched slightly negative median skewness ($\approx -0.3$), indicating that manipulation does not systematically skew global tone curves toward highlights or shadows.
    \item \textbf{Specular and Tail Discrepancies (Excess Kurtosis):} While both classes share a negative median kurtosis ($\approx -0.7$), authentic faces display a heavier distribution of positive outliers (extending up to $\approx 10$, compared to $\approx 3.5$ in fakes). Authentic captures contain sharp, localized specular reflections—such as corneal glints, wet lip specularities, and sharp directional lighting highlights—that generative autoencoders typically compress or oversmooth into diffused gradients.
\end{itemize}

Overall, the extensive interquartile overlap across all four moments demonstrates that global spatial intensity statistics alone cannot reliably distinguish authentic faces from manipulated ones. This confirms that forgery traces do not manifest as coarse global photometric shifts, substantiating the need for localized spatial-frequency representations and attention-based feature extraction.

\subsection{Radial Frequency Band Analysis on Isolated Faces}

\begin{figure*}[htbp]
    \centering
    \includegraphics[width=\linewidth]{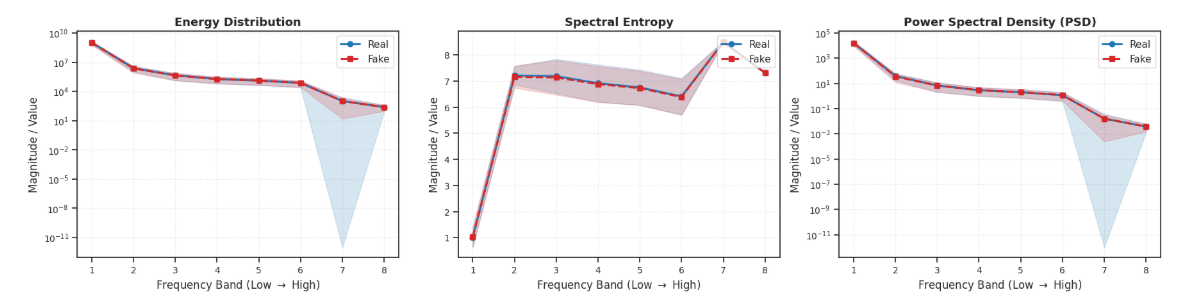}
    \caption{Empirical distributions (mean line with shaded $\pm 1\sigma$ standard deviation envelope) of Sub-band Energy, Spectral Entropy, and Power Spectral Density across 8 radial frequency bands computed exclusively over \texttt{dlib}-extracted facial crops for authentic (blue) and manipulated (red) images.}
    \label{fig:face_only_bands}
\end{figure*}

To investigate whether background scene clutter artificially influences frequency distributions, all facial regions are extracted and localized using a \texttt{dlib} frontal face detector with a 15\% contextual bounding margin prior to frequency analysis. Each isolated face is normalized to $256 \times 256$ pixels and converted to grayscale, after which a 2D Discrete Fourier Transform (DFT) with origin centering is computed:
\[
F_{\text{shifted}}(u, v) = \mathcal{F}\big\{I_{\text{face}}(x, y)\big\}
\]
The centered frequency spectrum is partitioned into $K=8$ concentric annular radial bands $\Omega_k$ ($k \in \{1, \dots, 8\}$). Across each radial band, we evaluate the dataset-wide mean and standard deviation ($\mu \pm 1\sigma$) for Sub-band Energy ($E_k$), Spectral Entropy ($H_k$), and Power Spectral Density ($S_k$) across all authentic and manipulated samples (Fig.~\ref{fig:face_only_bands}):
\[
E_k = \sum_{(u,v) \in \Omega_k} |F(u, v)|^2, \quad S_k = \frac{E_k}{M \cdot N}
\]
\[
H_k = -\sum_{(u,v) \in \Omega_k} P(u, v) \log\big(P(u, v) + \epsilon\big), \quad P(u, v) = \frac{|F(u, v)|^2}{E_k}
\]

Evaluating these metrics directly on cropped facial regions yields key forensic observations:
\begin{itemize}
    \item \textbf{Monotonic Spectral Attenuation:} Across both Energy and PSD, the metric curves exhibit a steep, power-law drop of over six orders of magnitude from Band~1 (low-frequency facial structure) to Band~8 (fine sensor grain and micro-texture), confirming that low-frequency macro-geometry numerically dominates the Fourier magnitude.
    \item \textbf{Global Cohort Overlap:} Despite removing all background pixels, the dataset-level mean trajectories for authentic and manipulated faces remain closely aligned across all three metrics. The standard deviation error envelopes ($\pm 1\sigma$) overlap heavily throughout Bands 1--6, demonstrating that intrinsic facial diversity (e.g., bone structure, expressions, and subject-specific skin tone) introduces more variance than the manipulation itself.
    \item \textbf{High-Frequency Divergence (Bands 7--8):} In the terminal radial bands (Bands 7 and 8), the authentic envelope maintains a higher upper margin than the synthesized cohort. This subtle separation captures natural camera sensor noise and uncompressed pores in authentic faces, which generative pipelines and autoencoder decoders partly smooth out.
    \item \textbf{Non-Monotonic Entropy Distribution:} Spectral entropy peaks sharply in intermediate bands (Bands 2--3) and terminal bands (Bands 7--8), reflecting the dispersion of facial edge transitions and noise. However, the scalar entropy distributions between the two classes remain largely indistinguishable.
\end{itemize}

These findings demonstrate that isolating the face from the background is insufficient for global scalar thresholding: global summary statistics aggregated over entire frequency bands obscure the localized tampering artifacts left by deepfake generators. This provides the primary motivation for maintaining full $H \times W$ spatial resolution in our multi-channel frequency tensor and applying Cross-Stream Attention Fusion (CSAF) to correlate localized frequency anomalies directly with spatial facial landmarks.

\subsection{Continuous Azimuthal Integration and High-Frequency Residual Analysis}

\begin{figure*}[htbp]
    \centering
    \includegraphics[width=\linewidth]{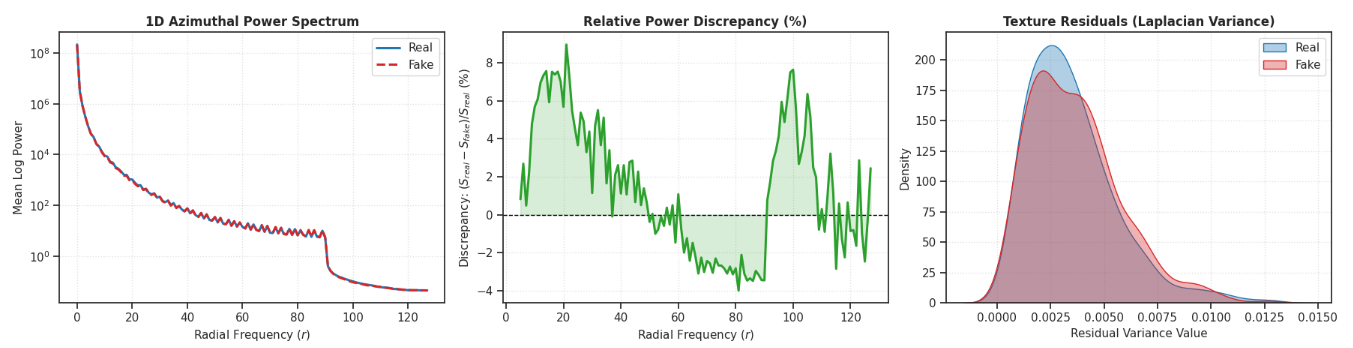}
     \caption{Continuous spectral and spatial residual analysis across authentic (blue) and manipulated (red) cohorts: (Left) 1D continuous azimuthal radial power spectrum $\bar{S}(r)$ on a logarithmic scale; (Middle) relative power discrepancy profile $\Delta S_{\text{rel}}(r) = \frac{\bar{S}_{\text{real}}(r) - \bar{S}_{\text{fake}}(r)}{\bar{S}_{\text{real}}(r)} \times 100\%$; (Right) kernel density estimation (KDE) of high-frequency spatial texture energy extracted via discrete Laplacian filtering.}
    \label{fig:continuous_spectral_analysis}
\end{figure*}

While partitioned radial bands demonstrate overall energy attenuation, discrete binning tends to smooth out localized frequency discrepancies across narrow spectral bands. To capture fine-grained spectral behavior without coarse discretization, we compute the continuous 1D azimuthal radial power spectrum via concentric circular integration from the origin to the Nyquist boundary ($r \in [0, M/2]$), alongside high-frequency spatial Laplacian residuals (Fig.~\ref{fig:continuous_spectral_analysis}):
\[
\bar{S}(r) = \frac{1}{|\mathcal{C}_r|} \sum_{(u,v) \in \mathcal{C}_r} |F_{\text{shifted}}(u, v)|^2, \] 
\[\mathcal{C}_r = \left\{(u,v) \;\middle|\; \left\lfloor \sqrt{u^2 + v^2} \right\rfloor = r\right\}
\]
where $\mathcal{C}_r$ denotes the set of frequency coordinates at radial distance $r$.

The resulting continuous spectral and residual distributions reveal distinct forensic signatures across three critical operational bands:
\begin{itemize}
    \item \textbf{Continuous Azimuthal Decay Profile:} As shown in the left panel, both authentic and manipulated faces follow natural $1/f^\alpha$ power-law decay across macroscopic structural scales ($r < 40$). A steep cliff occurs near $r \approx 90$, corresponding to standard video compression cutoffs and bilinear anti-aliasing filters.
    \item \textbf{Band-Selective Relative Power Discrepancies:} Evaluating the normalized percentage difference (middle panel) resolves anomalies that coarse scalar binning conceals. In the mid-frequency range ($r \in [10, 35]$), authentic images maintain an $8\%\text{--}9\%$ power surplus over synthesized faces, confirming that autoencoder decoders fail to preserve natural micro-gradients and subtle facial shading. Conversely, an inverted discrepancy occurs at high frequencies ($r \in [65, 90]$), where synthesized faces exhibit excess power (reaching $-4\%$) caused by periodic checkerboard ringing artifacts introduced during transposed convolution upsampling. A secondary peak at $r \in [95, 105]$ reflects the absence of authentic camera sensor noise in synthetic faces.
    \item \textbf{Laplacian Residual Energy Distribution:} The right panel plots the density of high-frequency spatial variance computed via convolution with a discrete Laplacian kernel $K_L$:
    \[
    R_L(x, y) = I_{\text{gray}} * K_L, \quad \sigma_L^2 = \frac{1}{M \cdot N} \sum_{x,y} |R_L(x, y)|^2
    \]
    The manipulated distribution exhibits an extended heavy tail ($\sigma_L^2 > 0.005$) absent in natural faces. This positive skew confirms that while facial interiors undergo generative smoothing, the stitching seams and blending boundaries introduce sharp, localized second-order derivative spikes.
\end{itemize}

These continuous and residual metrics show that manipulation traces are band-specific and spatially localized rather than globally uniform, validating our choice to feed multi-channel spatial-frequency representations directly into the cross-stream attention mechanism.

\subsection{Spatially-Resolved Frequency Band Reconstructions}

\begin{figure*}[htbp]
    \centering
    \includegraphics[width=\linewidth]{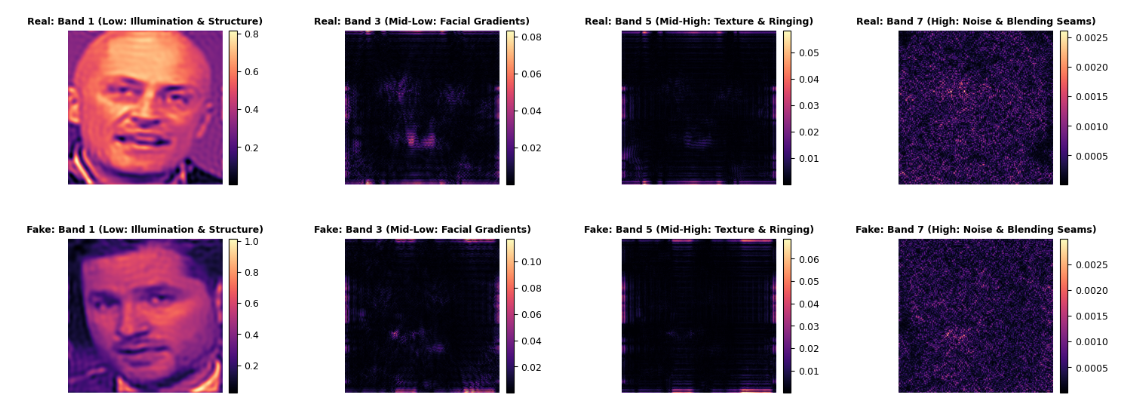}
   \caption{Multi-band spatial reconstructions via 2D Inverse DFT across representative radial bands (Bands 1, 3, 5, and 7): Row 1 displays authentic faces, and Row 2 displays manipulated faces.}
    \label{fig:spatial_reconstructions}
\end{figure*}

To examine how structural and textural facial information partitions across spatial scales, the input faces are transformed via 2D Discrete Fourier Transform (DFT), filtered into concentric annular radial bands, and reconstructed into the spatial domain ($H \times W$) via the 2D Inverse DFT (Fig.~\ref{fig:spatial_reconstructions}, Rows 1 and 2):
\[
I^{(k)}(x, y) = \left| \mathcal{F}^{-1} \left\{ F_{\text{shifted}}(u, v) \cdot \mathbb{I}_{\Omega_k}(u, v) \right\} \right|
\]
where $\Omega_k$ denotes the $k$-th annular band-pass mask.

Rows~1 and 2 highlight the progression of isolated facial features from low to high radial frequencies for authentic and manipulated faces, respectively:
\begin{itemize}
    \item \textbf{Band 1 (Low: Illumination \& Structure):} Reconstructions in the lowest radial band capture baseline facial morphology, global head contours, and macro-illumination gradients. However, this band is dominated by subject-specific identity characteristics rather than forgery signatures.
    \item \textbf{Band 3 (Mid-Low: Facial Gradients):} As the low-frequency DC offset is eliminated, coarse illumination disappears, isolating intermediate structural transitions such as cheek shading, nose bridge contours, and lip profiles. Natural facial gradients appear continuous in authentic faces, whereas deepfake faces exhibit subtle gradient flattening.
    \item \textbf{Band 5 (Mid-High: Texture \& Ringing):} Reconstructions in this intermediate band isolate fine facial boundaries, eyelid margins, and skin micro-textures. Synthesized faces in Row~2 show distinct periodic horizontal ringing striations and high-contrast boundary transitions left by transposed convolution upsampling and mask-blending algorithms.
    \item \textbf{Band 7 (High: Noise \& Blending Seams):} The highest radial band captures high-frequency edge transitions and camera sensor noise. While the authentic face displays a homogeneous grain distribution across the entire canvas, the manipulated face exhibits regional noise attenuation across inner facial regions paired with localized seam spikes along the boundary perimeter.
\end{itemize}

Preserving full $H \times W$ spatial resolution across these individual band reconstructions ensures that localized tampering artifacts—such as upsampling striations and boundary seams—remain spatially indexed for subsequent cross-stream attention layers rather than being discarded by global scalar averaging.

\subsection{Targeted High-Frequency Residual and Boundary Seam Analysis}

\begin{figure*}[htbp]
    \centering
    \includegraphics[width=\linewidth]{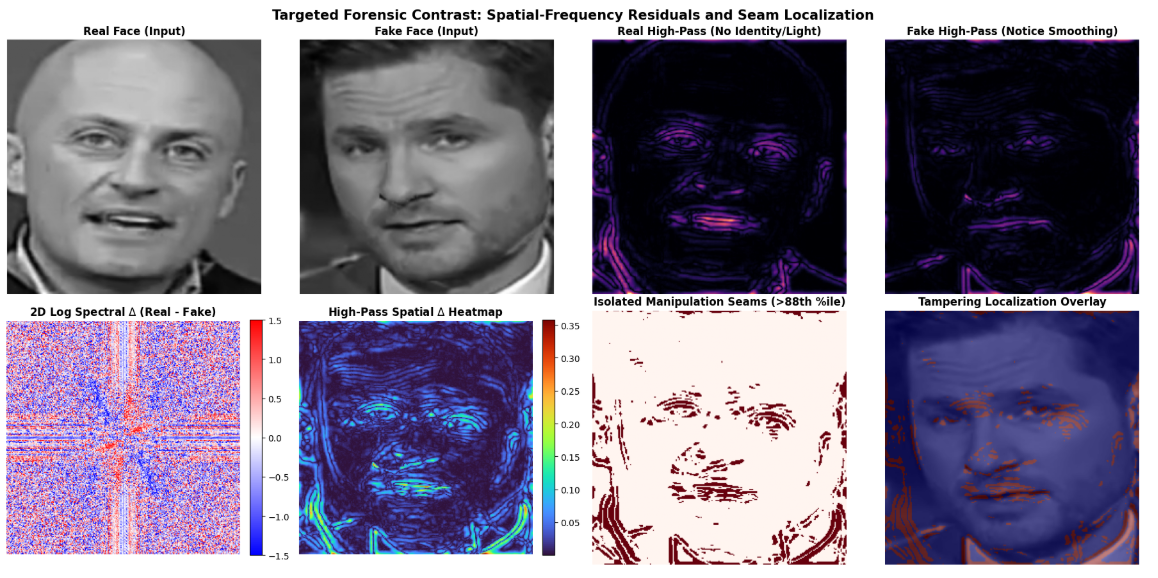}
    \caption{Targeted spatial-frequency forensics: (Top row, left-to-right) Cropped authentic input face, manipulated input face, Gaussian high-pass reconstructed authentic face, and high-pass reconstructed manipulated face; (Bottom row, left-to-right) 2D log-magnitude spectral difference ($\log |F_{\text{real}}| - \log |F_{\text{fake}}|$), spatial residual heatmap ($|I_{\text{HPF, real}} - I_{\text{HPF, fake}}|$), top 12\% thresholded anomaly seams, and the resulting tampering localization overlay.}
    \label{fig:targeted_forensics}
\end{figure*}

To eliminate the dominant influence of facial identity and broad illumination, we isolate high-frequency residuals and directional frequency discrepancies, as illustrated in Fig.~\ref{fig:targeted_forensics}. 

The top row compares the cropped grayscale inputs ($256 \times 256$) against their high-pass spatial reconstructions obtained via an inverse Fourier transform after suppressing the DC and low-frequency components ($r < 15$ pixels) using a circular Gaussian filter:
\[
H_{\text{HPF}}(u, v) = 1 - \exp\left(-\frac{u^2 + v^2}{2 \sigma_0^2}\right)
\]
This filtering removes broad lighting variations and facial morphology, isolating micro-textural patterns. While authentic skin exhibits dense, continuous sensor grain and fine pore structures, synthesized faces show distinct texture attenuation and oversmoothing caused by decoder interpolation.

The bottom row isolates the manipulation traces across both spectral and spatial dimensions:
\begin{itemize}
    \item \textbf{2D Log-Magnitude Spectral Discrepancy:} Computing $\log |F_{\text{real}}(u, v)| - \log |F_{\text{fake}}(u, v)|$ reveals non-isotropic deviations, highlighting checkerboard upsampling artifacts and frequency suppression that isotropic radial averaging obscures.
    \item \textbf{Spatial Residual Heatmap:} The pixel-wise absolute difference $|I_{\text{HPF, real}} - I_{\text{HPF, fake}}|$ demonstrates that residual errors are not uniformly distributed across the face; rather, they peak along high-frequency transition zones.
    \item \textbf{Seam Isolation and Tampering Localization:} Isolating residual outliers above the 88th percentile eliminates background noise and delineates localized tampering boundaries. When superimposed onto the target face, these outliers align with mask blending borders, eyelid contours, and perioral transitions.
\end{itemize}

This visual evidence confirms that deepfake manipulation signatures reside primarily in localized, high-frequency boundary seams rather than global scalar shifts, providing empirical motivation for our cross-stream spatial-frequency attention mechanism.
\section{Architecture}
First, the image is passed to the input layer, where it is resized and the pixel values are normalized. Next, it is sent to the preprocessing layer for frequency domain analysis, followed by spatial- and frequency-based encoders. The extracted features are passed to a cross-stream attention fusion layer, then to embedding and token refinement layers, and finally, a decision is taken to classify the image as real or fake.

\subsection{Input Layer}\label{AA}
The proposed model takes an RGB image as input, which is first pre-processed to ensure that it is ready for analysis. The image is resized to a fixed size, to maintain consistency across all inputs. The pixel values are then normalized by adjusting their range according to the mean and standard deviation values ($ x_i \rightarrow \frac{x_i - \mu}{\sigma}$), helping the model learn more effectively as the values are constrained to a standard range. Finally, the processed image is converted into a tensor format, which is passed to the next steps in the model pipeline.


\subsection{Preprocessing Layer}
To efficiently detect subtle manipulations present in deepfake images, the model utilizes frequency domain analysis. In contrast to the spatial domain, which analyzes pixel arrangements, the frequency domain reveals patterns in textures, edges, and inconsistencies that are often introduced during image manipulation (i.e., deepfake synthesis). The input image is transformed into the frequency domain using techniques such as the Discrete Fourier Transform (DFT). These methods break the image down into its frequency components, separating smooth, low-frequency regions (like backgrounds or skin tones) from sharp, high-frequency details (like edges and textures). 

In the proposed model, the image is first converted to grayscale to focus more on intensity—which predominantly reflects changes due to manipulation—and to reduce computational complexity.
\[
I_{\text{gray}}(x, y) = 0.2989 I_R(x, y) + 0.5870 I_G(x, y) + 0.1140 I_B(x, y)
\]

Then, Discrete Fourier Transform (DFT) is applied, converting the spatial features to frequencies and highlighting complex patterns:
\[
F(u, v) = \sum_{x=0}^{M-1} \sum_{y=0}^{N-1} I_{\text{gray}}(x, y) \cdot e^{-j 2\pi \left(\frac{ux}{M} + \frac{vy}{N}\right)}
\]
where \(F(u, v)\) is the Fourier coefficient at frequency \((u, v)\), \(I_{\text{gray}}(x, y)\) is the grayscale intensity at pixel \((x, y)\), and \(M, N\) are the dimensions of the image.

The zero-frequency component is then shifted to the center for easier analysis:
\[
F_{\text{shifted}}(u, v) = F\left((u + M/2) \mod M, (v + N/2) \mod N\right)
\]

The magnitude and phase of the frequency components are calculated, normalized, and stacked.
Magnitude:
\[
|F(u, v)| = \sqrt{\text{Re}(F(u, v))^2 + \text{Im}(F(u, v))^2}
\]
Phase:
\[
\phi(u, v) = \arctan\left(\frac{\text{Im}(F(u, v))}{\text{Re}(F(u, v))}\right)
\]
where \(\text{Re}(F(u, v))\) and \(\text{Im}(F(u, v))\) are the real and imaginary parts of the Fourier coefficient \(F(u, v)\).

Magnitude Normalization:
\[
|F_{\text{norm}}(u, v)| = \frac{|F(u, v)| - \min(|F(u, v)|)}{\max(|F(u, v)|) - \min(|F(u, v)|)}
\]
Stacking Magnitude and Phase:
\[
\text{Features}(u, v) = \begin{bmatrix} |F_{\text{norm}}(u, v)| \\ \phi(u, v) \end{bmatrix}
\]

The frequency spectrum is divided into multiple bands using band-pass filters, allowing the model to focus on specific ranges of frequencies. From these bands, important statistical features—such as energy distribution, entropy, and power spectral density (PSD)—are extracted to highlight critical patterns. 

Energy:
\[
E = \sum_{f_1}^{f_2} |X(f)|^2
\]
where \( X(f) \) is the Fourier transform of the signal at frequency \( f \), and \( |X(f)| \) is the magnitude of the frequency component. 

Entropy:
\[
H = - \sum_{f} P(f) \log(P(f))
\]
where \( P(f) = \frac{|X(f)|^2}{\sum_{f} |X(f)|^2} \) is the normalized power spectral density, representing the probability distribution of power at frequency \( f \). 

Power Spectral Density (PSD): 
\[ 
S(f) = \frac{|X(f)|^2}{T} 
\] 
where \( |X(f)|^2 \) is the squared magnitude of the Fourier transform \( X(f) \) at frequency \( f \), and \( T \) is the total duration of the signal.

These features are then restructured into feature maps that retain the image’s spatial resolution but include multiple channels, each representing a different frequency band. For instance, an image resized to \(224 \times 224\) might produce feature maps of size \(224 \times 224 \times 8\), where the 8 channels correspond to different frequency ranges.

\subsection{Dual-Stream Feature Encoder}

To efficiently capture both spatial details and frequency-based features, the dual-stream feature encoder processes the input through two parallel paths: the Spatial Feature Encoder and the Frequency Feature Encoder. These streams independently extract complementary representations, which are later fused through cross-attention to enhance the model’s ability to detect deepfake-specific anomalies.

\subsubsection{Spatial Feature Encoder}

The spatial feature encoder, a convolutional backbone without fully connected layers, takes in the original RGB image (\(H \times W \times C\), e.g., \(224 \times 224 \times 3\)). We tested with EfficientNet-B4 as the baseline convolutional model. These architectures are optimized for efficiency and accuracy in feature extraction. The backbone applies a series of convolutional layers with residual connections, which preserves low-level details while enabling deeper layers to focus on higher-level features.

Convolution Operation:
\[
f_{i,j,k} = \sum_{m,n} \sum_{c} I_{i+m, j+n, c} \cdot K_{m,n,c,k}
\]
where \(I_{i,j,c}\) is the input pixel value at position \((i, j)\) for channel \(c\), \(K_{m,n,c,k}\) is the kernel weight at position \((m, n)\) for channel \(c\) and filter \(k\), and \(f_{i,j,k}\) is the output feature map value at position \((i, j)\) for filter \(k\). Then, ReLU activation is applied:
\[
f'(x) = \max(0, x)
\]

Down-sampling is performed via pooling layers and stride-based convolutions to reduce the spatial resolution and increase the depth of the feature map. 
Pooling Operation (e.g., max pooling):
\[
P_{i,j,k} = \max_{m,n} f_{s \cdot i+m, s \cdot j+n, k}
\]
where \(s\) is the stride. A Global Average Pooling (GAP) layer is also used:
\[
GAP_k = \frac{1}{H \cdot W} \sum_{i=1}^H \sum_{j=1}^W f_{i,j,k}
\]

The output of this stream is a spatial feature map (\(S\)) of size \(P \times P \times C_2\), where \(P \times P\) represents the downsampled resolution, and \(C_2\) is the number of feature channels. In our case, using EfficientNet-B4 with the final fully connected layers removed, the output dimensions are \( 56 \times 56 \times 128 \).

\subsubsection{Frequency Feature Encoder} 

The frequency feature encoder processes the frequency-domain features obtained from the preprocessing layer, which have dimensions \(H \times W \times C\) (e.g., \(224 \times 224 \times 8\) in our case). This stream uses a shallow transformer module, utilizing multi-head self-attention to capture global dependencies and relationships in the frequency data. Specifically, a BERT like Smaller Sequential Encoder, which applies bidirectional processing, has been utilized for this stream.

The frequency feature map is divided into non-overlapping patches of size \(k \times k\), which are flattened into tokens, forming a sequence of \(N\) tokens.
\[
\text{Patch}_{i,j} = \{I_{m,n} \mid m \in [i \cdot k, (i+1) \cdot k), n \in [j \cdot k, (j+1) \cdot k)\}
\]
\[
t_p = W \cdot \text{Flatten}(\text{Patch}_{i,j}) + b
\]
where \(W\) is the projection matrix and \(b\) is the bias vector.

For an input of \(H \times H\) and patch size \(k \times k\), this results in \(N = (H/k)^2\) tokens. Positional embeddings are added to the tokens to retain spatial information.
\[
T_p = t_p + E_p
\]
where \(E_p\) is the positional embedding.

The sequence is then processed through a shallow transformer, where multi-head self-attention captures long-range dependencies and global patterns in the frequency domain.
\[
\text{Attention}(Q, K, V) = \text{softmax}\left(\frac{QK^\top}{\sqrt{d_k}}\right) V
\]
\[
\text{MHSA}(T_p) = \text{Concat}(\text{head}_1, \text{head}_2, \dots, \text{head}_h) W_O
\]
where \(W_O\) is the output projection matrix.

The output of this stream is a frequency embedding (\(F\)) of size \(P \times P \times C_3\) (e.g., \(56 \times 56 \times 128\)), where \(P \times P\) represents the reduced spatial resolution, and \(C_3\) is the number of embedding channels. A Feed-Forward Network (FFN) layer is also used:
\[
\text{FFN}(x) = \sigma(xW_1 + b_1)W_2 + b_2
\]
where \(W_1, W_2, b_1, b_2\) are learned parameters. The final frequency embedding is:
\[
F = \text{Reshape}(\text{Sequence Output})
\]
where \(F \in \mathbb{R}^{P \times P \times C_3}\).

The spatial encoder excels at identifying localized patterns such as edges, shapes, and textures, while the frequency encoder focuses on global patterns and artifacts commonly associated with manipulation. The outputs from these encoders, \(S\) and \(F\), are subsequently fused, enabling the model to leverage both spatial and frequency-based insights for robust deepfake detection. This dual-stream approach ensures that the model effectively analyzes both local and global inconsistencies introduced by image manipulation.

\subsection{Cross-Stream Attention Fusion (CSAF)}

The Cross-Stream Attention Fusion (CSAF) module integrates spatial and frequency feature maps to produce a unified representation, combining complementary insights from both streams. This fusion mechanism enhances the model's ability to detect deepfake-specific inconsistencies by aligning spatial details with frequency-domain artifacts.

The module receives two inputs: the Spatial Feature Map (\(S\)), output by the spatial feature encoder, with dimensions \(P \times P \times C_2\) (e.g., \(56 \times 56 \times 128\)), and the Frequency Embedding (\(F\)), output by the frequency feature encoder, with dimensions \(P \times P \times C_3\) (e.g., \(56 \times 56 \times 128\)).

The fusion process is driven by a cross-attention mechanism, where each stream's feature map is refined using complementary features from the other. First, the spatial feature map \(S\) is treated as the query, and the frequency embedding \(F\) serves as the key-value pair, enabling spatial features to attend to relevant frequency patterns. Similarly, frequency features are updated by attending to spatial information by interchanging the query and key-value pairs. The attention operation is mathematically represented as:
\[
\text{Attention}(Q, K, V) = \text{Softmax}\left(\frac{Q K^T}{\sqrt{d_k}}\right)V
\]
where \(Q\), \(K\), and \(V\) are the query, key, and value matrices derived from \(S\) and \(F\), and \(d_k\) is the scaling factor (dimensionality of \(K\)).

This bidirectional cross-attention ensures that spatial features guide the interpretation of frequency-based patterns and vice versa, leading to a robust, enriched representation.

The attention-enhanced feature maps from both streams are combined via summation. To refine the fused features, a feed-forward network is applied, ensuring the unified representation captures essential details from both spatial and frequency domains. Additionally, residual connections are incorporated to preserve the original feature information, with the final fused representation computed as:
\[H_1 = \text{CSAF}(S, F) + F\]
\[H_2 = \text{CSAF}(F, S) + S\]
\[H_{\text{fused}} = H_1 + H_2\]
where \(\text{CSAF}(S, F)\) denotes that \(S\) is treated as the query and \(F\) as the key-value pair.

The CSAF module produces a fused feature map \(H_{\text{fused}}\) with dimensions \(P \times P \times C_f\) (e.g., \(56 \times 56 \times 128\)), where \(C_f = C_2 = C_3\). This unified feature map encodes spatial details and frequency-domain artifacts in a cohesive manner, effectively capturing both localized manipulations and global inconsistencies introduced by deepfake generation techniques.

Spatial features provide fine-grained details, such as texture and edges, while frequency features capture global anomalies and artifacts. The cross-attention mechanism ensures effective communication between these streams, creating a unified representation that is well-suited for downstream tasks like classification and anomaly detection. This design ensures robustness against diverse manipulation techniques and improves overall model performance.

\begin{figure*}[htbp]
\centering
\includegraphics[height=30mm,width=.80\linewidth]{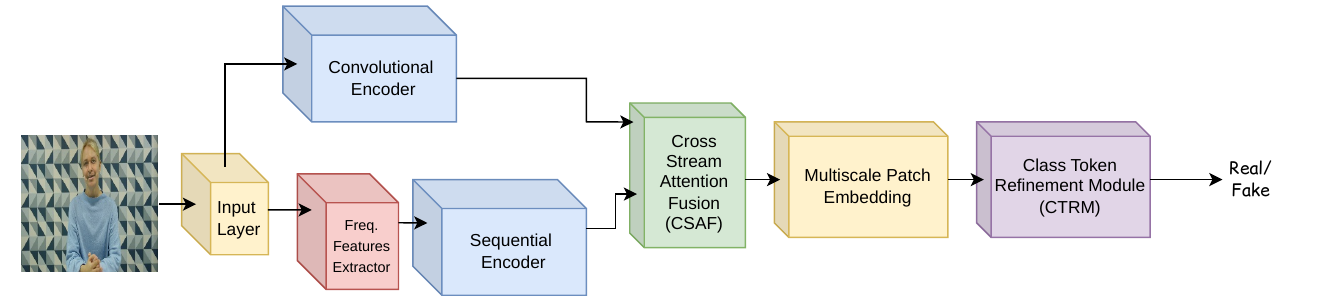}  
\caption{Our Proposed Architecture}
\label{fig:arch}
\end{figure*}


\subsection{Multiscale Patch Embedding}\label{SCM}

The Multiscale Patch Embedding module plays a vital role in capturing both fine details and broader patterns within the fused feature map (\(H_{\text{fused}}\)) produced by the Cross-Stream Attention Fusion (CSAF) module. This feature map, with dimensions \(P \times P \times C_f\) (e.g., \(56 \times 56 \times 128\)), is divided into patches of various sizes, such as \(k_1 \times k_1\), \(k_2 \times k_2\), and so on. Larger patches are used to focus on global structures and long-range dependencies, while smaller patches target localized and intricate details. Each patch is flattened into a token, with the number of tokens (\(T\)) at a given scale calculated as \(T = \left(\frac{P}{k}\right)^2\). To preserve spatial context, learnable positional embeddings are added to the tokens. These multi-scale tokens are then processed in parallel using lightweight transformers, each specialized in extracting features at its respective resolution. Finally, the outputs from all scales are fused through weighted summation, producing a unified multi-scale representation. The output of this layer is \((B, T_{\text{min}}, D)\), where \(B\) is the batch size, \(T_{\text{min}}\) corresponds to the smallest number of tokens across scales, and \(D = C_f\) is the embedding dimension. This design ensures that the module captures the full spectrum of spatial variations, from subtle texture mismatches to broad structural inconsistencies. By combining local details with global context, the multi-scale patch embedding significantly enhances the model's ability to detect deepfake manipulations effectively and reliably.

\subsection{Class Token Refinement Module (CTRM)}

The Class Token Refinement Module (CTRM) is the final processing step in the architecture, responsible for condensing the rich, multi-scale features into a compact representation for classification. At the heart of this module is a learnable class token (\( \mathbf{C} \in \mathbb{R}^{1 \times d_{\text{embedding}}} \)), which acts as a global aggregator. This token is prepended to the sequence of multi-scale feature tokens (\( \mathbf{F} \in \mathbb{R}^{T \times d_{\text{embedding}}} \), where \( T \) is the number of tokens and \( d_{\text{embedding}} \) is the embedding dimension) produced by the previous module, which can be represented as:
\[
\mathbf{X}_0 = \text{concat}(\mathbf{C}, \mathbf{F}) \in \mathbb{R}^{(T+1) \times d_{\text{embedding}}}
\]
This sequence is then processed through a transformer encoder, inside which the class token interacts with all other tokens using multi-head self-attention, allowing it to gather both localized and global information critical for deepfake detection.
\[
\text{Attention}(\mathbf{Q}, \mathbf{K}, \mathbf{V}) = \text{softmax}\left(\frac{\mathbf{Q} \mathbf{K}^\top}{\sqrt{d_k}}\right)\mathbf{V}
\]
where \( \mathbf{Q}, \mathbf{K}, \mathbf{V} \in \mathbb{R}^{(T+1) \times d_k} \) are the query, key, and value matrices derived from \( \mathbf{X}_l \) using learnable projection matrices, and \( d_k = \frac{d_{\text{embedding}}}{h} \) is the dimension per attention head (with \( h \) heads). Then, Multi-Head Aggregation is performed:
\[
\text{MultiHead}(\mathbf{X}_l) = \text{concat}(\text{head}_1, \dots, \text{head}_h)\mathbf{W}_o
\]
where \( \text{head}_i = \text{Attention}(\mathbf{Q}_i, \mathbf{K}_i, \mathbf{V}_i) \) and \( \mathbf{W}_o \in \mathbb{R}^{d_{\text{embedding}} \times d_{\text{embedding}}} \) is the output projection matrix.

Through this process, the class token refines its understanding of the input, distilling the most important features into a single representation. A Feed-Forward Network (FFN) further enhances the class token’s features, with residual connections and layer normalization ensuring stability and retaining essential information.
\[
\text{FFN}(\mathbf{X}_l) = \text{ReLU}(\mathbf{X}_l \mathbf{W}_1 + \mathbf{b}_1)\mathbf{W}_2 + \mathbf{b}_2
\]
where \( \mathbf{W}_1 \in \mathbb{R}^{d_{\text{embedding}} \times d_{\text{hidden}}} \), \( \mathbf{W}_2 \in \mathbb{R}^{d_{\text{hidden}} \times d_{\text{embedding}}} \), and \( d_{\text{hidden}} \) is the intermediate dimension in the FFN. 

Residual Connections and Layer Normalization are applied as follows:
\[
\mathbf{X}_{l+1}' = \text{LayerNorm}(\mathbf{X}_l + \text{MultiHead}(\mathbf{X}_l))
\]
\[
\mathbf{X}_{l+1} = \text{LayerNorm}(\mathbf{X}_{l+1}' + \text{FFN}(\mathbf{X}_{l+1}'))
\]

The result is a refined class token, with dimensions \(1 \times d_{\text{embedding}}\), that encapsulates the key characteristics of the input image. This representation is then passed to the final classification layer to determine whether the image is real or a deepfake. By focusing on the most critical aspects of the extracted features, the CTRM ensures that the model achieves high reliability and accuracy in detecting subtle manipulations.

\subsection{Final Classification Head}

The Final Classification Head is the last step of the architecture, where the refined class token from the Class Token Refinement Module (CTRM) is transformed into a meaningful prediction. This class token, a compact representation of the input's critical features, is passed through a fully connected layer that maps it to a single value, called the logit, for binary classification. A sigmoid activation function is then applied to the logit to generate a probability score between 0 and 1, indicating the likelihood of the input being a deepfake.
\[
\hat{y} = \text{sigmoid}(\mathbf{C}_{\text{refined}} \mathbf{W}_c + b_c)
\]
where \( \mathbf{W}_c \in \mathbb{R}^{d_{\text{embedding}} \times 1} \), \( b_c \in \mathbb{R} \), and \( \hat{y} \) is the predicted probability that the input is a deepfake. Let this probability be $p_i$. 

\section{Details of Dataset}
\textbf{FaceForensics++:} Rössler et al. introduced FaceForensics++, one of the most widely used benchmark datasets for face forgery detection, containing thousands of video sequences manipulated using four different methods: DeepFakes, Face2Face, FaceSwap, and NeuralTextures. The dataset provides videos at multiple compression levels, allowing researchers to evaluate detector robustness under realistic quality degradation.

\textbf{Celeb-DF:} Li et al. introduced Celeb-DF, a large-scale deepfake dataset built from real YouTube videos of celebrities, with corresponding deepfake versions generated using an improved synthesis algorithm designed to reduce visual artifacts present in earlier datasets. The improved generation pipeline addresses issues such as color inconsistency, temporal flickering, and low-resolution synthesized faces that made earlier deepfakes easy to detect. This higher visual quality makes Celeb-DF a more challenging benchmark than prior datasets, better reflecting the capabilities of modern deepfake generation methods. It has since become a standard testbed for evaluating the generalization ability of detection models.

\textbf{WildDeepfake:} Zi et al. introduced WildDeepfake, a deepfake detection dataset collected entirely from real-world internet sources rather than being generated under controlled laboratory conditions. The videos include diverse scenes, lighting conditions, resolutions, and manipulation methods, making the dataset substantially more challenging than earlier, more curated benchmarks. The authors argue that this real-world diversity better reflects the conditions under which deepfake detectors must operate in practice.

\textbf{Deep Fake Detection (DFD) Dataset:} This Kaggle-hosted dataset provides the original and entire Deep Fake Detection (DFD) dataset, a collection of real and manipulated face videos released as part of a broader effort to support deepfake detection research. It contains face-swapped videos generated using multiple actors under varied conditions, alongside the corresponding source footage. The dataset is commonly used alongside other benchmarks such as FaceForensics++ to test model generalization across different generation sources. Its inclusion in evaluation protocols helps assess whether a detector's performance is specific to one dataset or generalizes across independently produced forgeries.

\textbf{The Deepfake Detection Challenge (DFDC) Preview Dataset:} Dolhansky et al. introduce the preview dataset for the Deepfake Detection Challenge, a large-scale collection of videos featuring paid actors whose faces were swapped using multiple different deepfake generation methods. The dataset was designed with careful attention to diversity in subjects, lighting, and pose to encourage the development of detectors that generalize beyond a narrow set of conditions. This preview release preceded the full DFDC dataset and associated Kaggle competition, which became one of the largest and most influential deepfake detection benchmarks. The paper also discusses the data collection methodology and licensing considerations involved in creating a dataset with consenting participants.



\section{Implementation Details}
The images are resized to a spatial dimension of $224\times224$. The pre-processing step involves detecting key facial landmarks and cropping the image to focus on the face. Random JPEG compression, Gaussian noise, motion blur, color jittering, and down-sampling are applied during training. This ensures that the model operates on relevant facial features, improving the accuracy of the attention mechanism. For details, we have followed the standard pre-processing of Deepfake-Bench~\cite{b43}.  The batch size is fixed at 32 for all experiments. In the training module, we employ the Adam optimization algorithm with a learning rate of 0.0002.  We sample 32 frames for each video for training and testing.

\section{Results}
The model has been tested on various deepfake detection datasets, such as FaceForensics++ (FF++) \cite{b37}, Celeb-DF (CDF) \cite{b38}, WildDeepfake (WDF) \cite{b39}, DeepFakeDetection (DFD) \cite{b41}, and the DeepFake Detection Challenge (DFDC) \cite{b40} datasets, which contain both real and fake images or videos. 

Our proposed architecture (Ours) utilizes EfficientNet-B4 and BERT-like Sequential encoder. The results are summarized in Table~\ref{tab:model_comparison_intra} and Table~\ref{tab:model_comparison_cross}. The tables show that our method achieves results competitive with the SOTA, specifically 99.55\% and 99.38\% AUC on FF++ and Celeb-DF, respectively. The approach generalizes well, achieving strong cross-dataset results. When trained on the FF++ dataset, the method achieves 83.28\% AUC on Celeb-DF, 65.11\% on WDF, 69.27\% on DFDC, and 91.23\% on DFD datasets.

\begin{table}[htbp]
\caption{Comparison with SOTA on Intra-Dataset Analysis. The Metric used is Video Level AUC.}
\centering
\begin{tabular}{lcc}
\toprule
\textbf{Model} & \multicolumn{2}{c}{\textbf{AUC (\%)} } \\
\cmidrule(lr){2-3}
 &  \textbf{FF++} & \textbf{Celeb-DF} \\
\midrule
Xception \cite{b25} & 96.30 & 99.73 \\
EfficientNet-B4 \cite{b26}  & 99.70 & 99.81 \\
Multi-Att \cite{b21}  & 99.29 & 99.94 \\
SPSL \cite{b28}  & 96.91 & - \\ 
RECCE \cite{b29} & 99.32 & 99.94 \\ 
FAce-X-Ray \cite{b30}  & 99.17 & - \\ 
LRL \cite{b31}  & 99.46 & - \\ 
SBIs (EfficientNet-B4) \cite{b32}  & 99.64 & 93.74 \\ 
SBIs (Swin Transformer) \cite{b32} & 99.72 & 95.68 \\ 
\midrule
\textbf{Ours}  & 99.55 & 99.38 \\
\bottomrule
\end{tabular}
\label{tab:model_comparison_intra}
\end{table}

\begin{table}[!t]
\caption{Cross-Domain Testing, Trained on FF++ c23 and tested on these Unseen Datasets (Video Level AUC (\%))}
\centering
\begin{tabular}{lcccc}
\toprule
\textbf{Model} & \textbf{CDF} & \textbf{WDF} & \textbf{DFDC} & \textbf{DFD} \\
\midrule
Xception \cite{b25} & 61.80 & 62.72 & 48.98 & 87.86 \\
EfficientNet-B4 \cite{b26}  & 64.29 & 63.83 & - & - \\
Multi-Att \cite{b21}  & 67.44 & 59.74 & - & - \\
SPSL \cite{b28}  & 76.88 & - & 66.16 & - \\ 
RECCE \cite{b29} & 68.71 & 64.31 & 69.06 & - \\ 
FAce-X-Ray \cite{b30}  & 80.58 & - & 80.92 & 95.40 \\ 
LRL \cite{b31}  & 78.26 & - & 76.53 & 89.24 \\ 
SBIs (EfficientNet-B4) \cite{b32}  & 93.18 & - & 72.42 & 97.56 \\ 
SBIs (Swin Transformer) \cite{b32} & 89.12 & 70.56 & 71.08 & 97.34 \\ 
\midrule
\textbf{Ours} & 83.28 & 65.11 & 69.27 & 91.23 \\
\bottomrule
\end{tabular}

\vspace{0.5em}
\small{Feature Extracting Backbone (Ours): EfficientNet-B4}
\label{tab:model_comparison_cross}
\end{table}
\section{Conclusion}
In this paper, a unified and novel approach to multi-modal deepfake detection was introduced. The approach integrates spatial and frequency domain features through cross-attention, followed by multi-scale patch embedding and a class token refinement module. Our method achieves results competitive with the SOTA, specifically 99.55\% and 99.38\% AUC on FF++ and Celeb-DF, respectively, using EfficientNet-B4 and BERT-like Sequenatial Encoder. The approach generalizes well, demonstrating competitive cross-dataset results.

\bibliographystyle{IEEEtran}
\bibliography{main} 

\end{document}